# Evolving Boolean Networks with RNA Editing

Larry Bull

Department of Computer Science & Creative Technologies
University of the West of England, Bristol BS16 1QY, UK
Larry.Bull@uwe.ac.uk

**Abstract**

The editing of transcribed RNA by other molecules such that the form of the final product differs from that specified in the corresponding DNA sequence is ubiquitous. This paper uses an abstract, tunable Boolean genetic regulatory network model to explore aspects of RNA editing. In particular, it is shown how dynamically altering expressed sequences via a guide RNA-inspired mechanism can be selected for by simulated evolution under various single and multicellular scenarios.

## Introduction

RNA editing is the alteration of the sequence of RNA molecules through a variety of mechanisms after initial expression (see (Maas, 2013) for an overview). In some cases such editing is triggered by specific conditions, in others it is necessary for the normal function of a cell. RNA editing is widespread and appears to have evolved many times (eg, see (Gray, 2012)). For example, guide RNA (gRNA) are relatively small molecules that align themselves to complementary regions of messenger RNA (mRNA) and either insert or delete a bases(s) thereby (typically) altering the structure of the protein specified in the expressed DNA.

With the aim of enabling the systematic exploration of artificial genetic regulatory network models (GRN), a simple approach to combining them with abstract fitness landscapes has been presented (Bull, 2012). More specifically, random Boolean networks (RBN) (Kauffman, 1969) were combined with the NK model of fitness landscapes (Kauffman & Levin, 1987). In the combined form – termed the RBNK model – a simple relationship between the states of $N$ randomly assigned nodes within an RBN is assumed such that their value is used within a given NK fitness landscape of trait dependencies. The approach was also extended to enable consideration of coevolutionary and multicellular scenarios using the related NKCS landscapes (Kaufmann & Johnsen, 1991) – termed the RBNKCS model. In this paper, RBNs are extended to include a simple form of RNA editing. The selection of the extra mechanism is explored under various single and multiple cell scenarios. Results indicate RNA editing is useful across a wide range of conditions. The paper is arranged are follows: the next section briefly reviews related work in the area and introduces the two basic models; the next section examines the extended RBNK model; and, the following examines the extended RBNKCS model. Finally, all findings are discussed.

## Background

### RNA Editing

There appears to be very little work which considers the role of RNA editing explicitly. After (Rocha, 1995), Huang et al. (see (2007) for an overview) have explored the use of a stochastic template matching mechanism which either inserts or deletes binary genes for function optimization. They report consistent benefit for dynamic/non-stationary functions in particular. Rohlfshagen and Bullinaria (2006) used an RNA editing-inspired scheme as a repair function for multi-constrained knapsack problems. Some formal models of aspects of RNA editing have also been presented (eg, (Liu & Bundschuh, 2005)). No previous work with artificial GRN is known.

### The RBNK Model

Within the traditional form of RBN, a network of $R$ nodes, each with a randomly assigned Boolean update function and $B$ directed connections randomly assigned from other nodes in the network, all update synchronously based upon the current state of those $B$ nodes. Hence those $B$ nodes are seen to have a regulatory effect upon the given node, specified by the given Boolean function attributed to it. Since they have a finite number of possible states and they are deterministic, such networks eventually fall into an attractor. It is well-established that the value of $B$ affects the emergent behaviour of RBN wherein attractors typically contain an increasing number of states with increasing $B$ (see (Kauffman, 1993) for an overview). Three regimes of behaviour exist: ordered when $B=1$, with attractors consisting of one or a few states; chaotic when $B \geq 3$, with a very large number of states per attractor; and, a critical regime around $B=2$, where similar states lie on trajectories that tend to neither diverge nor converge. Note that traditionally the size of an RBN is labelled $N$, as opposed to $R$ here, and the degree of node connectivity labelled $K$, as opposed to $B$ here. The change is adopted due to the traditional use of the labels $N$ and $K$ in the NK model of fitness landscapes which are also used in this paper, as will be shown.

Kauffman and Levin (1987) introduced the NK model to allow the systematic study of various aspects of fitness landscapes (see (Kauffman, 1993) for an overview). In the standard NK model an individual is represented by a set of $N$ (binary) genes or traits, each of which depends upon its own value and that of $K$ randomly chosen others in the individual. Thus increasing $K$, with respect to $N$, increases the epistasis. This increases the ruggedness of the fitness landscapes by increasing the number of fitness peaks. The NK model assumes all epistatic interactions are so complex that it is only appropriate to assign (uniform) random values to their effects on fitness. Therefore for each of the possible $K$ interactions, a table of $2^{(K+1)}$ fitnesses is created, with all entries in the range 0.0 to 1.0, such that there is one fitness value for each combination of traits. The fitness contribution of each trait is found from its individual table. These fitnesses are then summed and normalised by $N$ to give the selective fitness of the individual. Three general classes exist: unimodal when $K=0$; uncorrelated, multi-peaked when $K>3$; and, a critical regime around $0<K<4$, where multiple peaks are correlated.

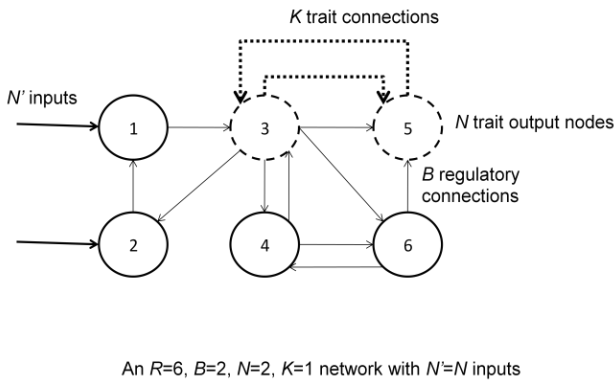

Figure 1. Example RBNK model with an equal number of input and output nodes. Dashed lines and nodes indicate where the NK fitness landscape is embedded into the RBN.

As shown in Figure 1, in the RBNK model $N$ nodes (where $R≤N<0$) in the RBN are chosen as outputs/traits, i.e., their state determines fitness using the NK model. The combination of the RBN and NK model enables a systematic exploration of the relationship between phenotypic traits and the genetic regulatory network by which they are produced. It was previously shown how achievable fitness decreases with increasing $B$, how increasing $N$ with respect to $R$ decreases achievable fitness, and how $R$ can be decreased without detriment to achievable fitness for low $B$ (Bull, 2012). In this paper $N$ phenotypic traits are attributed to randomly chosen nodes within the network of $R$ genetic loci, with environmental inputs applied to the first $N'$ loci (Figure 1); input nodes and trait/output nodes are not necessarily disjoint. Hence the NK element creates a tunable component to the overall fitness landscape with behaviour (potentially) influenced by the environment. For simplicity, $N'=N$ here.

**The RBNKCS Model**

Kauffman highlighted that species do not evolve independently of their ecological partners and subsequently presented a coevolutionary version of the NK model. Here each node/gene is coupled to $K$ others locally and to $C$ (also randomly chosen) within each of the $S$ other species/individuals with which it interacts – the NKCS model (Kauffman & Johnsen, 1991). Therefore for each of the possible $K+C$x$S$ interactions, a table of $2^{(K+1+C\text{x}S)}$ fitnesses is created, with all entries in the range 0.0 to 1.0, such that there is one fitness value for each combination of traits. The fitness contribution of each gene is found from its individual table. These fitnesses are then summed and normalised by $N$ to give the selective fitness of the total genome (see (Kauffman, 1993) for an overview). It is shown that as $C$ increases, mean fitness drops and the time taken to reach an equilibrium point increases, along with an associated decrease in the equilibrium fitness level. That is, adaptive moves made by one partner deform the fitness landscape of its partner(s), with increasing effect for increasing $C$. As in the NK model, it is again assumed all intergenome ($C$) and intragenome ($K$) interactions are so complex that it is only appropriate to assign random values to their effects on fitness.

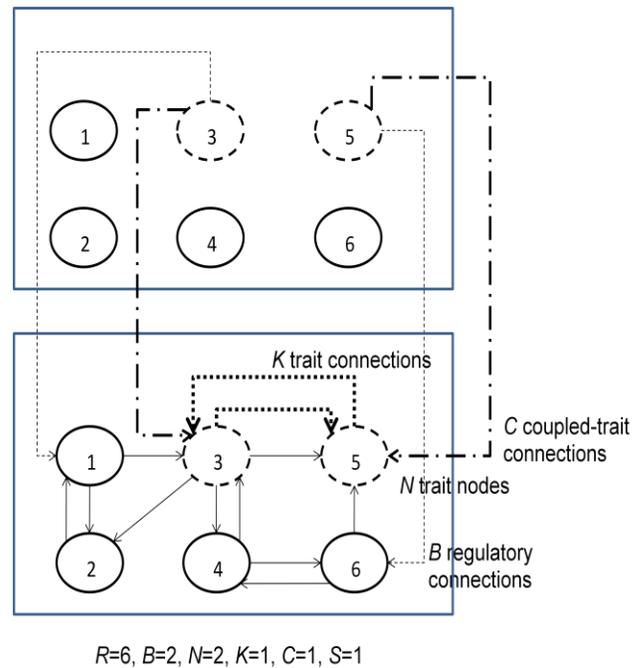

Figure 2. Example RBNKCS model. Connections for only one of the two coupled networks are shown for clarity.

The RBNK model is easily extended to consider the interaction between multiple GRN based on the NKCS model – the RBNKCS model. As Figure 2 shows, it is here assumed that the current state of the $N$ trait nodes of one network provide input to a set of $N$ internal nodes in each of its coupled partners, i.e., each serving as one of their $B$ connections. Similarly, the fitness contribution of the $N$ trait nodes considers not only the $K$ local connections but also the $C$ connections to its $S$ coupled partners' trait nodes. The GRN update alternately.

# RNA Editing in the RBNK(CS) Model

## gRNA

To include a mechanism which enables the modification of transcription based upon the internal and/or external environment of the cell, nodes in the RBN can (potentially) include a second set of $B'$ connections to defined nodes and another Boolean update function. Each such node also maintains a table containing a list of node id's for each entry in the Boolean table for the $B'$ connections where the node is on/expressed. The list is the same size as the out-degree of the node (range $[0, R*B]$). This is seen as introducing a non-coding RNA associated with the protein expressed by the given node. RNA editing causes a change in the connectivity of the RBN which lasts for one update cycle.

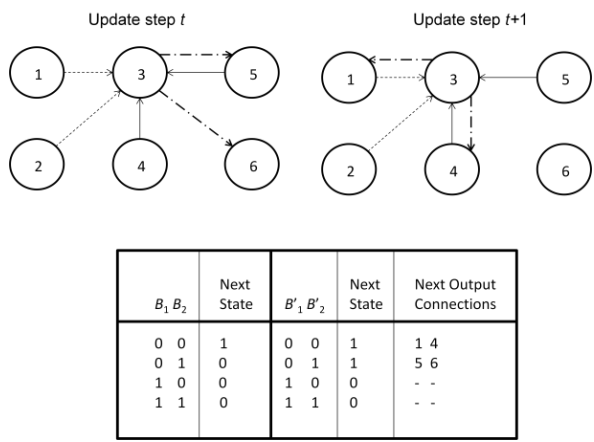

Figure 3. Example RBN with RNA editing. The look-up table and connections for node 3 only are shown for clarity. If every node in the RBN is in state '0' at time 't', node 3 will turn on for the next time step, as will its associated gRNA. As a consequence, nodes 1 and 4 will be connected to node 3 for time step $t$+1 for their updating.

On each traditional RBN update cycle, the connectivity of the network is initially assumed (reset) to be that originally determined by evolution. Then, for each node which has an associated guide RNA, a check is made to see if the node's transcription state was set to on ('1') on the last update step and if its associated RNA has been activated ('1') since the last time this occurred. If so, the out-connections for that node are altered to those in the corresponding table entry for the current state of the $B'$ connection nodes (Figure 3). If a node subsequently has fewer than $B$ input connections, the missing gene(s) is simply assumed to not be expressed on that cycle, i.e., the gene on the end of the connection is assumed to be set to 0. If a node subsequently has more than $B$ connections, the "extra" input is randomly assigned to one of the existing $B$ connections and a logical OR function is used to determine whether that connection is considered to be to an expressed gene. Thereafter each node updates its transcription state based upon the current state of the nodes it is (currently) connected to using the Boolean logic function assigned to it in the standard way, as do any associated non-coding RNA node elements. For simplicity, the number of standard regulatory connections is assumed to be the same as for RNA editing, i.e., $B=B'$.

## RBNK Experimentation

For simplicity with respect to the underlying evolutionary search process, a genetic hill-climber is considered here, as in (Bull, 2012). Each RBN is represented as a list to define each node's start state, Boolean function for transcription, $B$ connection ids, $B'$ connection ids, Boolean function for RNA editing, re-connectivity entries under RNA editing, and whether it is an RNA edited node or not. Mutation can therefore either (with equal probability): alter the Boolean transcription function of a randomly chosen node; alter a randomly chosen $B$ connection; alter a node start state; turn a node into or out of being RNA editable; alter one of the re-connection entries if it is an editable node; or, alter a randomly chosen $B'$ connection, again only if it is an editable node. A single fitness evaluation of a given GRN is ascertained by updating each node for 100 cycles from the genome defined start states. An input string of $N'$ 0's is applied on every cycle here. At each update cycle, the value of each of the $N$ trait nodes in the GRN is used to calculate fitness on the given NK landscape. The final fitness assigned to the GRN is the average over 100 such updates here. A mutated GRN becomes the parent for the next generation if its fitness is higher than that of the original. In the case of fitness ties the number of RNA editable nodes is considered, with the smaller number favoured, the decision being arbitrary upon a further tie. *Hence there is a slight selective pressure against RNA editing*. Here $R$=100, $N$=10 and results are averaged over 100 runs - 10 runs on each of 10 landscapes per parameter configuration - for 50,000 generations, $0<B\leq5$ and $0\leq K\leq5$ are used. As Figure 4 shows, regardless of $K$, RNA editing is selected for in all high connectivity cases on average, i.e., when $B>3$, when the underlying fitness landscape is unchanging. Analysis of the behaviour of the editing in such cases indicates that it is applied throughout the lifecycle, although a clear a pattern of usage is typically difficult to establish, often with varying numbers of nodes exhibiting editing per cycle.

Since it is known such highly connected networks typically exhibit chaotic dynamics (Kauffman, 1993) and they are subsequently difficult to evolve (Bull, 2012), it might be surmised that the RNA editing is not performing a functional role, rather it is maintained under drift/neutral processes. As noted above, RNA editing alters the out-connections of a given node and hence a potential consequence is the alteration in the number of connections into a given node. In particular, given the seemingly positive selection of editing in the high $B$ cases in Figure 4 it might be assumed that the mechanism's ability to effectively reduce a node's $B$ is all that is being selected for since fitness drops with increasing $B$. Experiments (not shown) in which the out-connection table entries are randomly re-created in the offspring indicate a significant (T-test, $p<0.05$) drop in fitness in all cases where editing is selected for and hence evolution does appear to be shaping suitable, dynamic behaviour through the editing mechanism. Although some editing nodes are also almost certainly there due to drift/neutral processes.

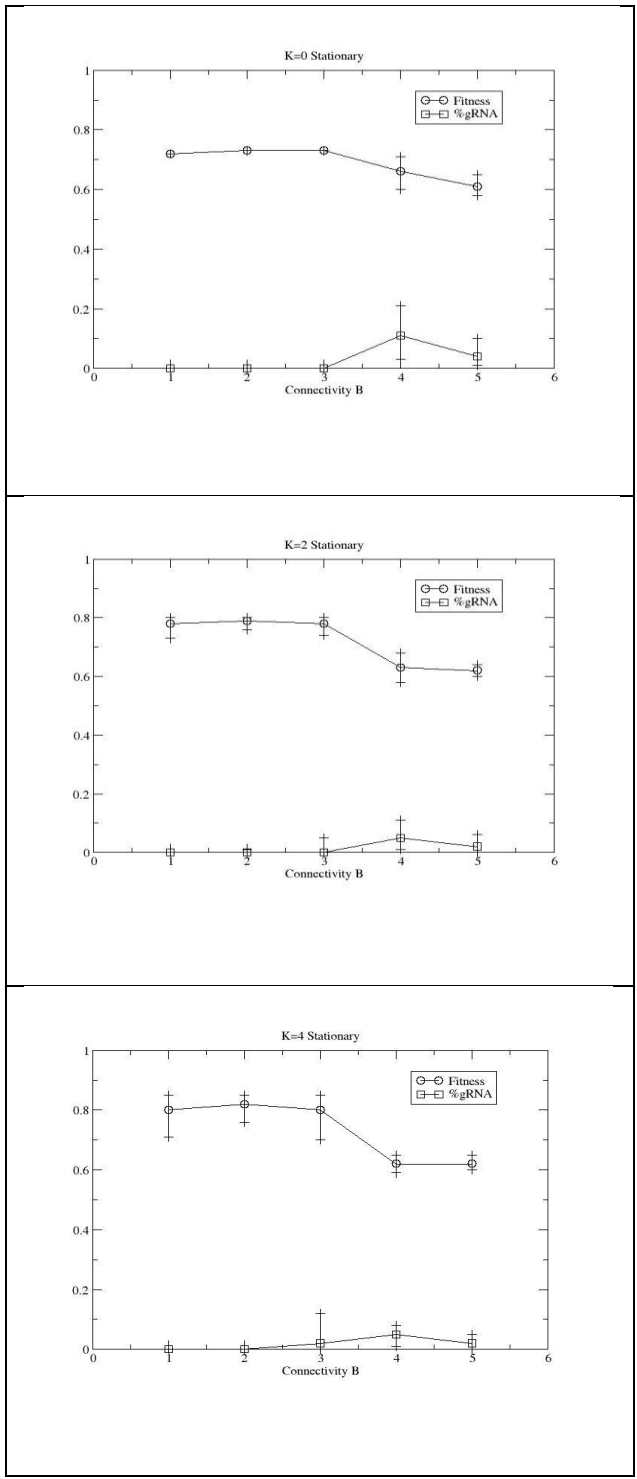

Figure 4. Evolutionary performance of RBN augmented with an RNA editing mechanism, after 50,000 generations. The percentage of nodes which use RNA editing ("%gRNA") is scaled 0-1, as is fitness. Error bars show min and max values.

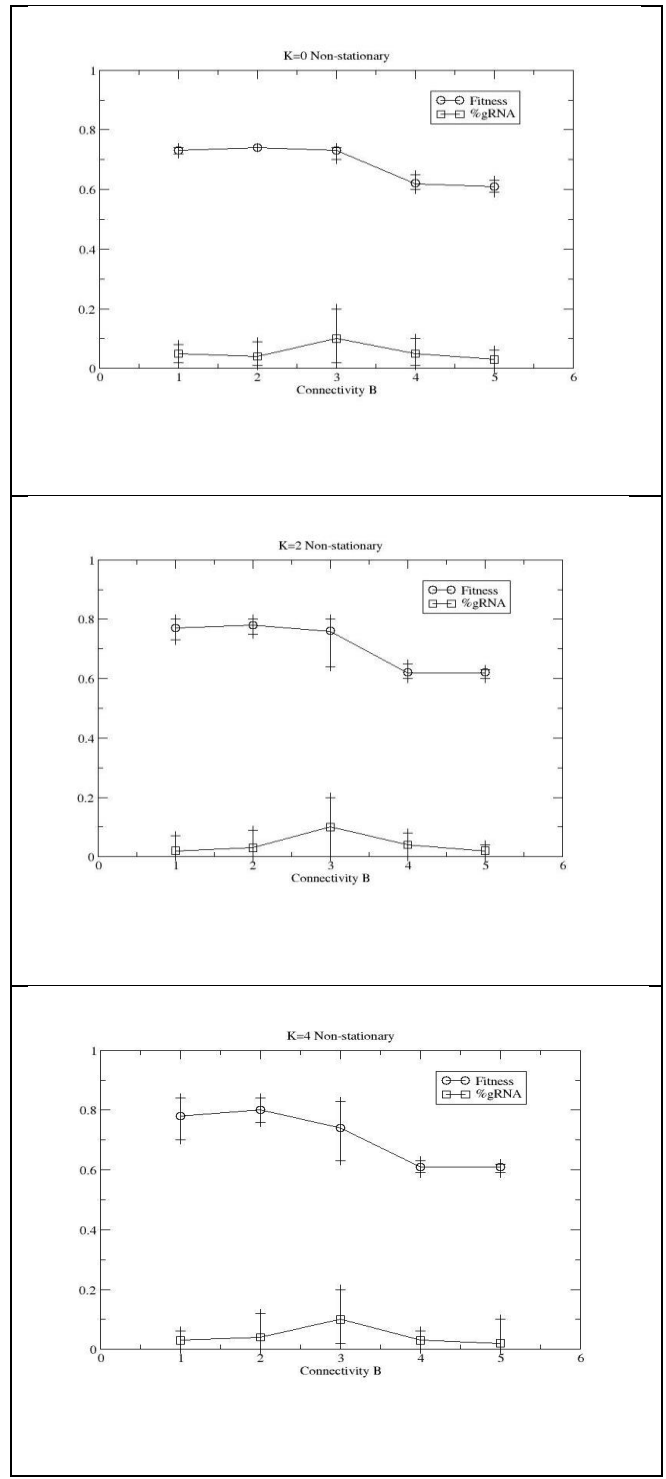

Figure 5. Evolutionary performance of RBN augmented with an RNA editing mechanism, after 50,000 generations, on a non-stationary fitness landscape. The percentage of nodes which use RNA editing ("%gRNA") is scaled 0-1, as is fitness. Error bars show max and min values.

Following (Huang et al., 2007), Figure 5 shows how RNA editing is selected for under all conditions when the underlying fitness landscape changes halfway through the lifecycle; an input of all 1's is applied on update cycle 50 and fitness contributions are calculated over a second NK landscape. Analysis of typical behaviour in the low $B$ cases shows that one or two nodes use editing either up to or after the point of change. That is, the editing is used to make small changes to the network topology to compensate for the disruption in the environment; the RNA editing has an active (context sensitive) role in the cyclic behaviour of the networks. These general results were also found for other values of $R$, eg, $R$=200 (not shown).

## RBNKCS Experimentation

**Heterogeneous cells.** The case of two coevolving GRN has previously been explored using the RBNKCS model, each evolved separately on their own NKCS fitness landscape for their $N$ external traits (Bull, 2012). Each network updates in turn for 100 cycles. The fitness of one network is then ascertained and an evolutionary generation for that network is undertaken. The mutated network is evaluated with the same partner as the original and it becomes the parent under the same criteria as used above. Then the second species network is evaluated with that network, before a mutated form is created and evaluated against the same partner. One generation is said to have occurred when all four steps have been undertaken. Only the fitness of the species with the potential to exploit RNA editing is shown here. This general scenario is potentially of interest given the proposed role of RNA editing by cells against viruses (Grivell, 1993), for example.

Figure 6 shows results for a low degree of coupling between the two species/cells, ie, C=1. As before, low $B$ networks result in higher fitness levels. However, regardless of $K$, the percentage of nodes using RNA editing increases with $B$. For $B$ = 1 editing is not selected for on average. Recall that such networks typically exhibit a point or small attractor, and hence the coupled GRNs exist in relatively static/unchanging environments. For $B$>3 the majority of nodes use RNA editing (>60%) but the fitness levels reached are relatively low. As above, the same experiments have been run in which the RNA editing details are randomly scrambled in offspring. Results (not shown) indicate similar high percentages of RNA editing nodes but no significant change in fitness (T-test, $p$>0.05) and hence the uptake is due to drift/neutral processes within such poorly evolving systems. The exact ways in which the editing is used in the low $B$ cases is hard to establish. Figure 7 shows how the same general trends occur for higher levels of coupling between the two, i.e., $C$=5. It can be noted that some level of RNA editing is now seen when $B$=1.

Analysis of how the percentage of RNA editing nodes varies over time shows relatively stable behaviour in the RBNK model (not shown). Figure 8 shows example runs of how this is not the case in the RBNKCS model, rather the percentage varies over time. Indeed, it appears there is a rough correlation between periods of coevolutionary stasis with regards to fitness and a decrease in the percentage of RNA editing nodes, and vice versa, for lower values of $C$.

**Homogeneous cells.** RNA editing is known to play a role in cell differentiation, eg, region-specific editing in mammalian brains for serotonin receptors (Burns et al., 1997). The case of two interacting cells has previously been explored with the RBNKCS model, where one is the daughter (clone) of the other, ie, $S$=1 again. Following (Bull, 2014), the (reproducing) mother cell is updated through one cycle and then both update in turn for 100 cycles, thereby introducing some asymmetry in GRN states into the model, with the mother receiving the average fitness of the two cells. All other aspects remain the same as before, with each cell existing on a different fitness landscape; differentiation is assumed.

Figure 9 shows examples of how the degree of adoption of the RNA editing in such cases is almost identical to that seen in the non-stationary single cell cases above. That is, it is seen for all $B$ at relatively low levels. Similar results (not shown) are seen when task differentiation is not assume between the cells. Again, understanding how the mechanism is used in the coupled cells is non-trivial.

## Discussion

RNA editing appears to serve many roles (Gray, 2012), including (organelle) mutation repair and defence against viruses. From a regulatory network perspective, editing such as that provided by guide RNA enables context sensitive structural dynamism. This (temporary) re-wiring of the underlying network topology provides a further layer of regulation. The results in this paper suggest that the mechanism will be selected for across a wide variety of conditions, particularly non-stationary and multiple celled scenarios. Results also indicate selection in high connectivity cases. It can be noted that within natural GRN genes have low connectivity on average (eg, (Leclerc, 2008)), as RBN predict, but a number of high connectivity "hub" genes perform significant roles (eg, see (Barabási & Oltvai, 2004)). The results here suggest RNA editing may help facilitate such structures.

Future work should consider the use of RNA editing within other GRN representations and explore whether these results also hold for asynchronous updating schemes.

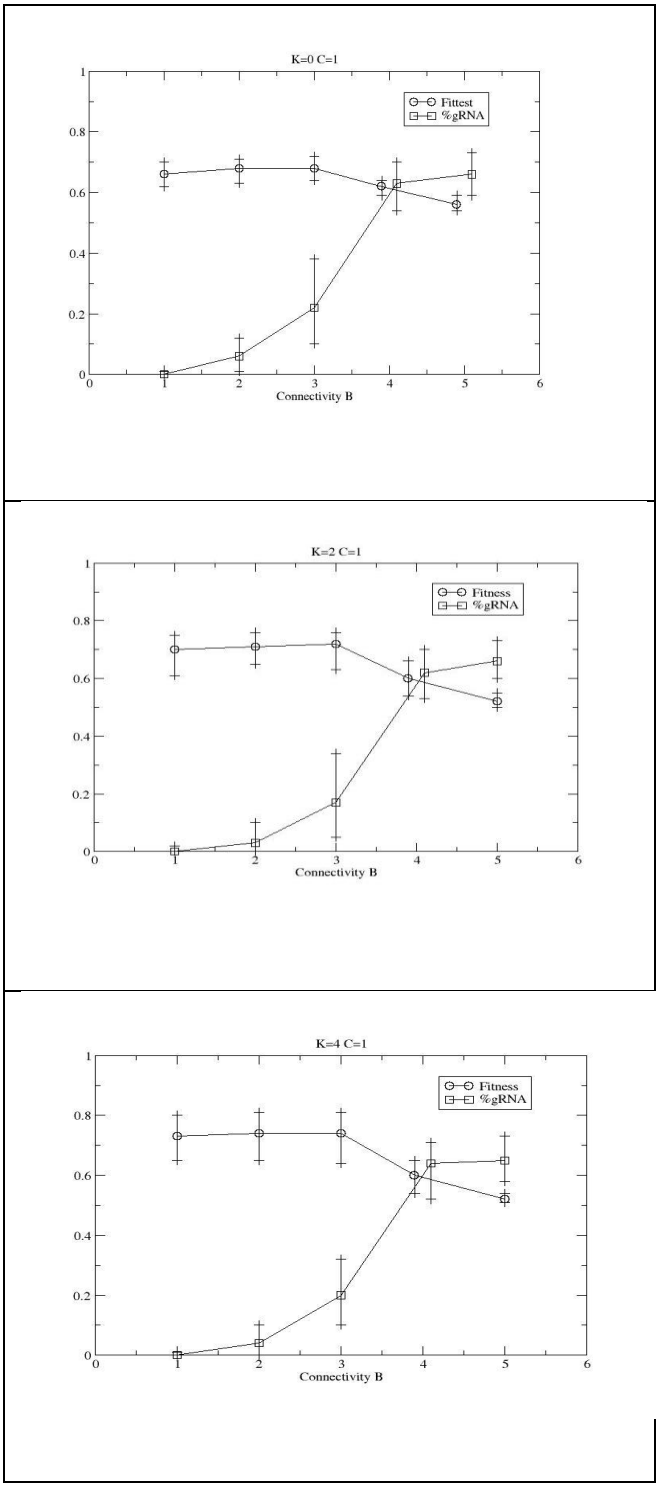

Figure 6. Performance of the augmented RBN coevolved against another, after 50,000 generations. The percentage of nodes which use RNa editing ("%gRNA") is scaled 0-1, as is fitness. Error bars show max and min values.

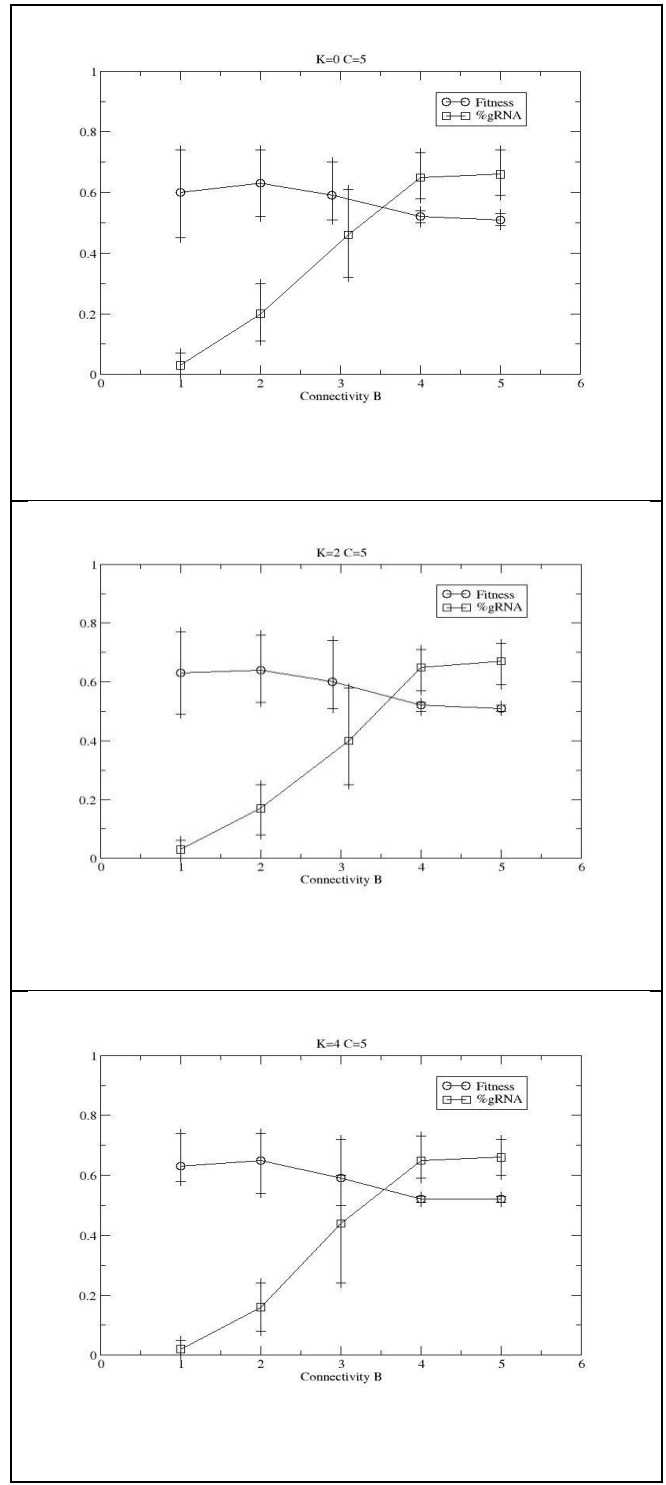

Figure 7. Performance of the augmented RBN coevolved against another, after 50,000 generations, for a high degree of coupling (*C*). The percentage of nodes which use RNA editing ("%gRNA") is scaled 0-1, as is fitness. Error bars show max and min values.

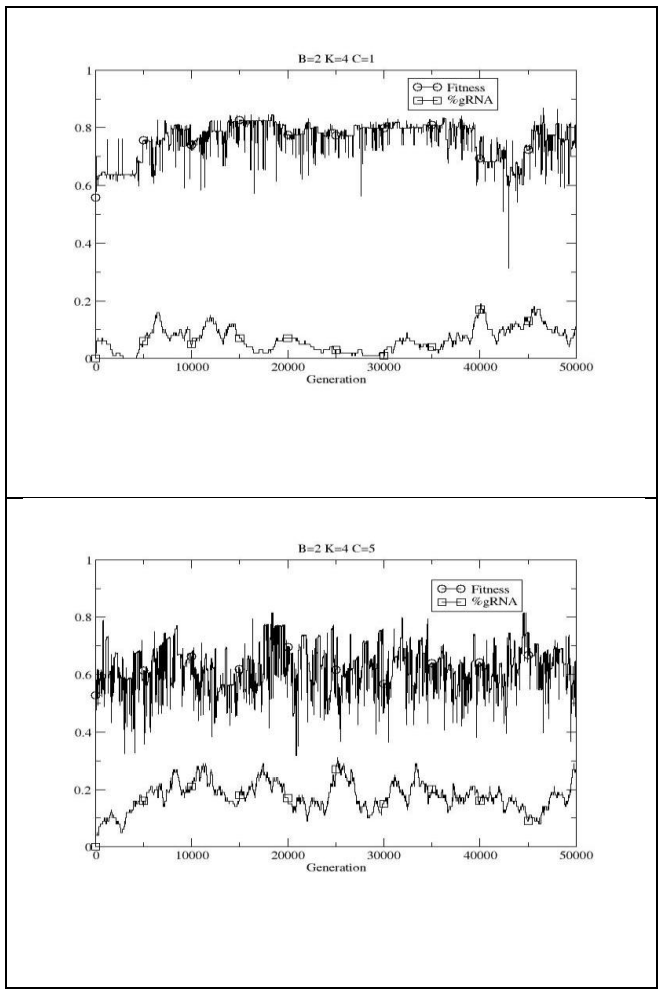

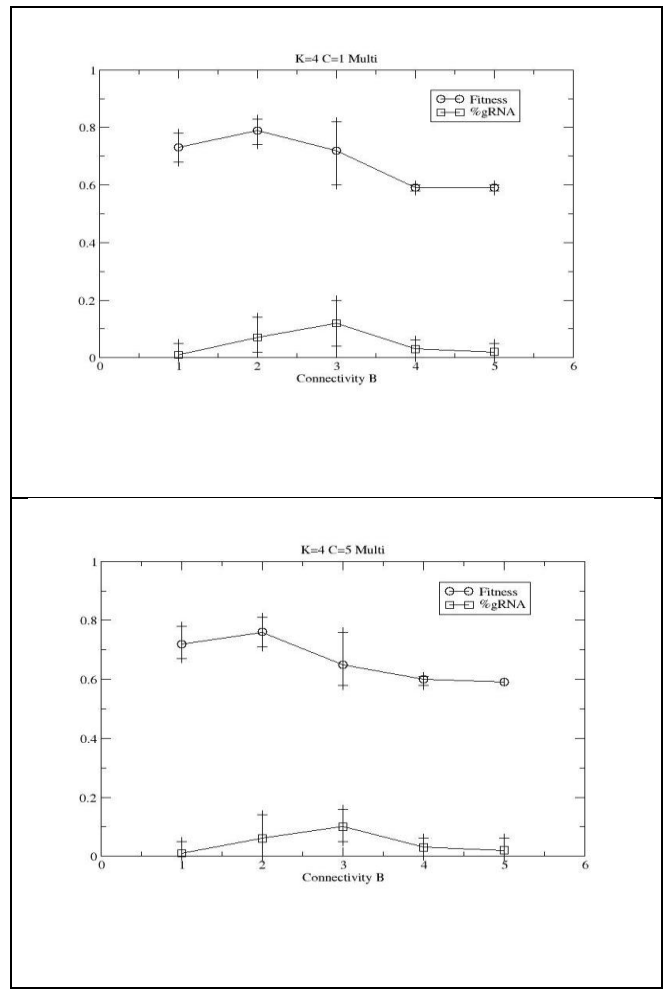

Figure 8. Example single runs of the coevolutionary case showing how editing is exploited to varying degrees depending upon the overall temporal dynamics of the ecosystem, rising during periods of re-adaptation and falling during periods of stasis for low (*C*). The percentage of nodes which use editing ("%gRNA") scaled as before.

Figure 9. Example performance in the two-celled case, after 50,000 generations. The percentage of nodes which use RNA editing ("%gRNA") scaled as before.